\documentclass[10pt,twocolumn,letterpaper]{article}

\usepackage{wacv}              

\usepackage{graphicx}
\usepackage{amsmath}
\usepackage{amssymb}
\usepackage{booktabs}
\usepackage{multirow}
\usepackage{bbm}

\usepackage[pagebackref,breaklinks,colorlinks]{hyperref}
\usepackage[accsupp]{axessibility}  

\usepackage[capitalize]{cleveref}
\crefname{section}{Sec.}{Secs.}
\Crefname{section}{Section}{Sections}
\Crefname{table}{Table}{Tables}
\crefname{table}{Tab.}{Tabs.}

\newcommand*{\affaddr}[1]{#1} 
\newcommand*{\affmark}[1][*]{\textsuperscript{#1}}
\newcommand*{\email}[1]{\texttt{#1}}

\def\wacvPaperID{2119} 
\def\confName{WACV}
\def\confYear{2024}

\begin{document}

\title{Lightweight Delivery Detection on Doorbell Cameras}

\author{%
Pirazh Khorramshahi\affmark[1]\thanks{\small{This work was done as the internship project at Comcast Applied AI Research.}}, \hspace{0.05cm} Zhe Wu\affmark[2], Tianchen Wang\affmark[2], Luke Deluccia\affmark[2] and Hongcheng Wang\affmark[3]\\
\affaddr{\affmark[1]Qualcomm Technologies},
\affaddr{\affmark[2]Comcast Applied AI Research},
\affaddr{\affmark[3]Amazon}\\
\email{\tt\small pirazhkhorramshahi@gmail.com, \{zhe\_wu, tianchen\_wang, luke\_deLuccia\}@comcast.com}, \\ \email{\tt\small hongcheng.wang@gmail.com} 
}

\maketitle

\begin{abstract}
Despite recent advances in video-based action recognition and robust spatio-temporal modeling, most of the proposed approaches rely on the abundance of computational resources to afford running huge and computation-intensive convolutional or transformer-based neural networks to obtain satisfactory results. This limits the deployment of such models on edge devices with limited power and computing resources. In this work we investigate an important smart home application, video based delivery detection, and present a simple and lightweight pipeline for this task that can run on resource-constrained doorbell cameras. Our method relies on motion cues to generate a set of coarse activity proposals followed by their classification with a mobile-friendly 3DCNN network. To train we design a novel semi-supervised attention module that helps the network to learn robust spatio-temporal features and adopt an evidence-based optimization objective that allows for quantifying the uncertainty of predictions made by the network. Experimental results on our curated delivery dataset shows the significant effectiveness of our pipeline and highlights the benefits of our training phase novelties to achieve free and considerable inference-time performance gains. 
\end{abstract}
\section{Introduction}
\label{sec:introduction}
\begin{figure}[t!]
    \centering
    \subfloat{\includegraphics[width=.11\textwidth, height=.11\textwidth]{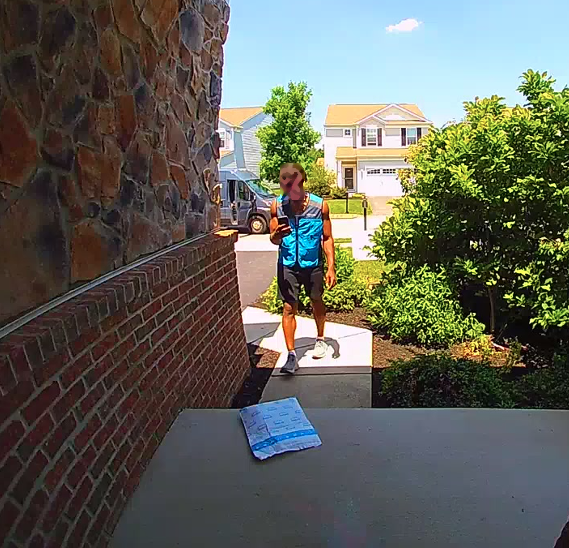}}~
    \subfloat{\includegraphics[width=0.11\textwidth, height=.11\textwidth]{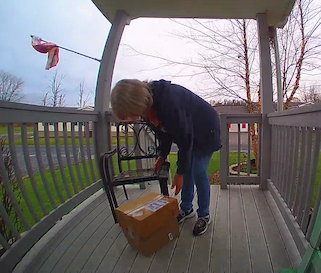}}~
    \subfloat{\includegraphics[width=0.11\textwidth, height=.11\textwidth]{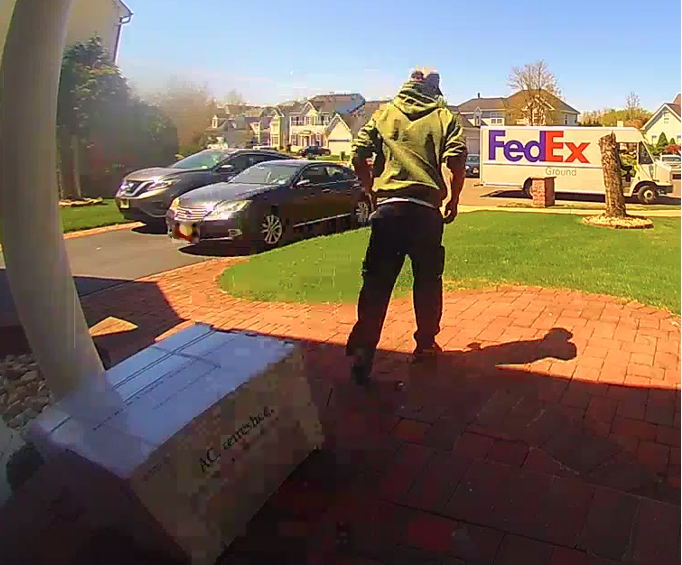}}~
    \subfloat{\includegraphics[width=0.11\textwidth, height=.11\textwidth]{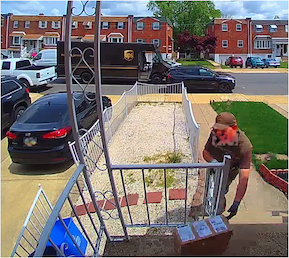}}
    \caption{Sample delivery events captured by doorbell cameras.}
    \label{fig:sample_deliveries}
\end{figure}

Computer Vision has become potent thanks to advances in Deep Learning to an extent that long standing problems like object detection, semantic segmentation, face and human action recognition can now be solved with high accuracy. Despite this, we have to highlight that this often comes at the price of significant computational burden which is typically accelerated by the use of Graphical Processing Units (GPU), Tensor Processing Units (TPU), or Neural Processing Units (NPU). Therefore, the degree to which computer vision tasks can be solved on edge devices with limited resources and computational power is constrained. Among these tasks is \textbf{Delivery Detection} which is concerned with recognizing delivery of merchandises (package, food, groceries, mail, etc.) at front doors to provide timely notifications for customers. Note that delivery detection task is different from package detection in that it identifies the instances of delivering items rather than the mere detection of packages which is currently practiced in smart home solutions. Delivery detection has numerous advantages including prevention of food perishing and porch piracy to name a few. According to the package theft annual report\footnote{{\scriptsize\url{https://security.org/package-theft/annual-report/}}}, in a twelve month period from 2021 to 2022, there has been more than 49 million package theft incidents in the United States alone with the estimated value of $\$2.4$B. The prevalence of smart devices including smart doorbell and security cameras through out residential locations facilitates the development and adoption of automated delivery detection systems which can significantly reduce losses. Fig. \ref{fig:sample_deliveries} shows captured deliveries by doorbell cameras.  
Despite potential applications, delivery detection is a challenging task. Type, shape and size of packages can be quite diverse. Cardboard boxes in various size, mail, grocery bags, and food are among the items that are frequently delivered. Additionally, there are various courier services including United States Postal Services (USPS), United Parcel Services (UPS), DHL and Amazon, as well as growing number of smaller companies like DoorDash and Uber Eats especially after the COVID-19 pandemic. This translates to delivery personnel having diverse outfits and appearances as evidenced by Fig. \ref{fig:sample_deliveries}. Finally the temporal extent of delivery events have high variance. Smaller items are delivered in a matter of seconds while delivering heavier objects can take much longer in the order of minutes. This also depends on submitting the proof of delivery in the form of a picture. Existing solutions such as Ring, Nest Hello, Arlo, AWS Rekognition, and Vivint mainly require cloud processing which results in higher bandwidth utilization, computation, and increased subscription fees. In addition, transferring data creates privacy concerns as opposed to local processing. Moreover, these methods primarily focus on detecting packages/boxes and not the instances of deliveries. In addition, package detection may fail as small or occluded packages are harder to be detected. Therefore, we set out to propose a solution to detect delivery instances that can be implemented on edge devices. We present a lightweight system that relies on motion detection to generate event proposals followed by their classification with a mobile-friendly 3DCNN \cite{ji20123d} network. Through the novel incorporation of an attention mechanism and benefiting from the the theory of evidence and subjective logic, we significantly improve the base performance of this system without imposing additional processing costs. In summary, this paper makes the following contributions:
\begin{itemize}
    \item We introduce a lightweight delivery detection system running on doorbell cameras with ARM Cortex-A family of processors. In contrast to the widely used package detection in the industry, our system is to detect the delivery events from videos.    
    \item We propose a semi-supervised attention module in 3DCNNs to extract robust spatio-temporal features.
    \item We propose to adopt evidential learning objective to quantify the uncertainty of predictions and enforce a minimum certainty score to ensure quality predictions.
\end{itemize}
\section{Related Work}
\label{sec:related_work}
Development of CNNs has substantially contributed to the remarkable improvements in the status of video action recognition. \cite{simonyan2014two} presented a two-stream design to leverage both RGB and Optical Flow modalities to capture spatio-temporal cues. With the prevalence of 3D convolutional kernels \cite{ji20123d}, I3D network was introduced to better model temporal interactions and established a strong baseline \cite{carreira2017quo}. Authors in \cite{gleason2019proposal} proposed a two-step approach to localize potential activities from hierarchical clustering of detected objects and recognize a wide range of activities in surveillance cameras from Optical Flow modality using a modified I3D, namely TRI3D, to adjust the temporal bounds of localized activities. Despite strong performance, I3D incurs high computational cost due to its depth and significant number of 3D filters. To reduce the computational burden, authors in \cite{xie2018rethinking} proposed S3D network in which only deeper layers of the network are designed to capture temporal information, namely top-heavy design. Additionally they propose to factorize 3D convolutional filters into spatial and temporal layers to reduce computational complexity. This is also proposed by \cite{tran2018closer} in their R(2+1)D network design to improve the efficiency of action recognition models. In another line of work \cite{feichtenhofer2019slowfast} proposed a two-path design to capture spatial semantics at a reduced frame rate and temporal cues at finer resolution in a faster and lighter pathway; the design known as slow-fast, has improved the accuracy/efficiency trade-off significantly. With the introduction of transformers \cite{vaswani2017attention}, many works soon adopted transformers in the context of video understanding. \cite{girdhar2019video} uses the base of I3D network to obtain initial spatio-temporal features upon which proposals are generated and corresponding features are passed to a stack of action transformer units. To improve efficiency, \cite{fan2021multiscale} proposed a multi-scale network MViT to generate multi-scale feature hierarchies by spatio-temporal down-sampling as well as down-sampling of the dimensionality in attention heads. Similarly, \cite{liu2022video} adopted SWIN transformer \cite{liu2021swin} for video action recognition to reduce the quadratic complexity of standard transformer modules. Authors in \cite{arnab2021vivit} designed ViViT with factorized encoding scheme to ingest tokenized input sequences and lower the complexity associated with the running of transformer blocks. Despite these progress, these models are heavyweight which limits their deployment on a device with limited power and compute budget \cite{koot2021evaluating}. There have been attempts to leverage transformer-based architectures for mobile applications; however, they are mainly limited to 2D vision applications \cite{mehta2021mobilevit} as scaling these models to 3D and temporal modeling is non-trivial. Other attempts to adopt transformers for human action recognition in a mobile environment, use non-visual modalities such as inertial measures \cite{ek2022lightweight} which cannot be used for spatially fixed cameras \emph{e.g.} surveillance cameras. Therefore, in this work, we present a lightweight CNN-based pipeline that can run on edge devices and consistently improve its performance by devising a novel attention module and benefiting from the the theory of evidence and subjective logic in the training objective.
\section{Method} 
\label{sec:method}
To the best of our knowledge, there are no published research on the delivery detection task. Therefore, we first establish a simple, intuitive, and easy to implement baseline. Next, we propose our novel delivery detection system to process untrimmed videos which overcomes the shortcomings of the baseline.

\subsection{Baseline System}
\label{sec:current_del_system}
\begin{figure}
    \centering
    \includegraphics[width=0.45\textwidth]{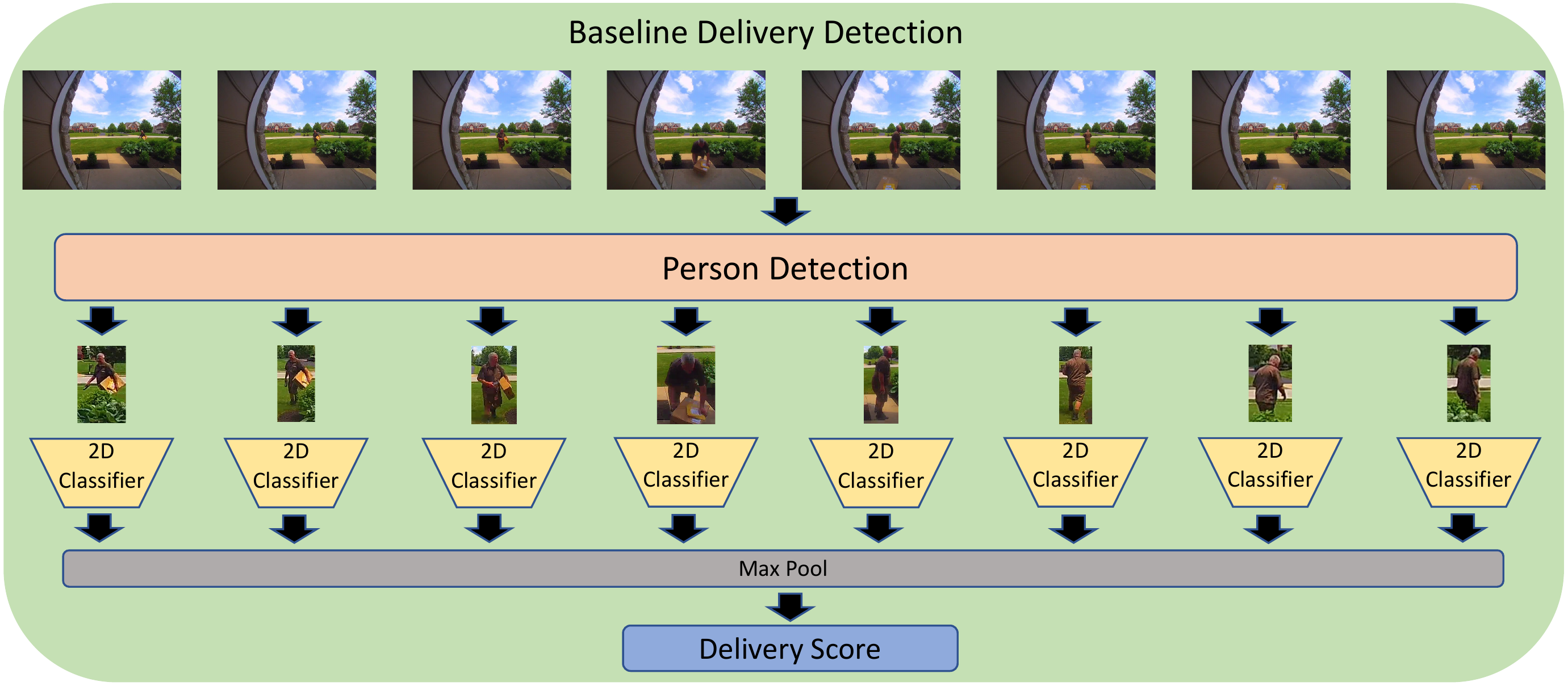}
    \caption{Baseline delivery detection pipeline. In each frame, a person detector localizes a person with highest detection confidence and passes the person's crop to a 2D classifier to obtain a delivery score. Delivery scores from all the sampled frames in the video snippet are max-pooled to generate the final delivery score.}
    \label{fig:baseline_pipeline}
\end{figure}

Our baseline model which is shown in Fig. \ref{fig:baseline_pipeline}, is a two-stage method where the first stage is responsible to identify whether a person is present in the scene for a given frame. In case a person is detected, second stage crops the person from the full frame, 
and passes it to a 2D classifier to generate a delivery score $s_i \in [0,1]$ ($i$ is the frame index) where larger $s_i$ indicates higher chance of a delivery personnel. The system queries the scene at a fixed rate of $1$ frame per second over the continuous intervals of length $15$ seconds. Therefore, each chunk is summarized by $15$ delivery scores; max-pooling is used to obtain the final delivery score, \emph{i.e.} $s = \max_{i=1}^{15} s_i$. Backbone architectures for person detector and 2D classifier are MobileNet-SSD \cite{liu2016ssd} and EfficientNet-B0 \cite{tan2019efficientnet} respectively to meet the constraints imposed by the resource-limited hardware of a doorbell camera. Adoption of max-pooling facilitates the implementation and results in an efficient system which can be easily interpreted. However, this cultivates a high False Positive Rate (FPR). In case, the second-stage outputs a relatively high delivery score for a single frame due to an artifact, or sudden variation in illumination, the system makes a false detection despite opposing evidence from other frames. This will be discussed in more details in section \ref{sec:experiments}. Moreover, this design requires two individual modules which prevents the end-to-end optimization of the overall system and the temporal information that can provide critical cues delivery recognition is not considered. This motivates us to devise a system that can model temporal interactions and is optimized in an end-to-end fashion. 
\subsection{Proposed Delivery Detection System}\label{sec:proposed_del_system}
\begin{figure}
    \centering
    \includegraphics[width=0.45\textwidth]{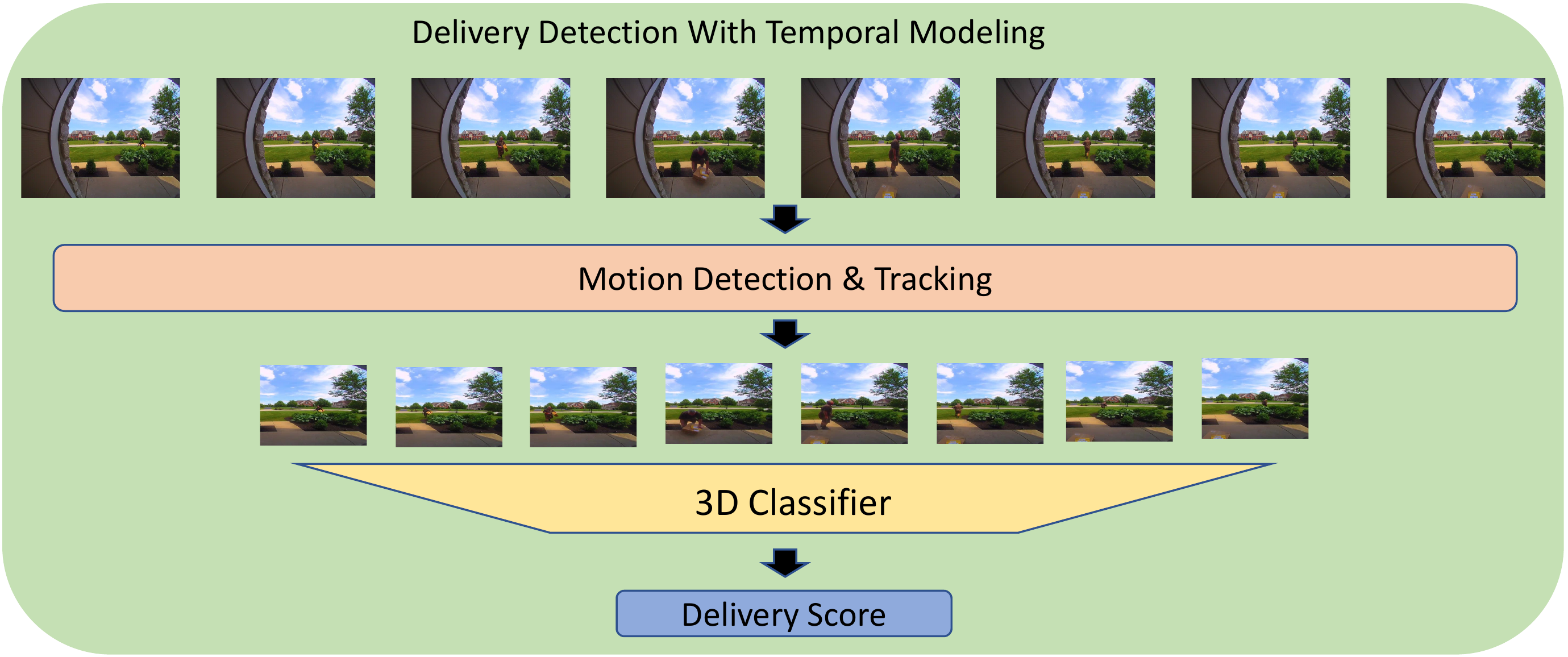}
    \caption{Proposed delivery detection pipeline. Motion algorithm detects and tracks foreground motion blobs. The foreground motion is used to reduce the spatial extent of frames to motion regions. Once certain number of frames are gathered, they are passed to a 3D classifier to obtain the delivery score.}
    \label{fig:proposed_pipeline}
\end{figure}

As mentioned above, rich temporal information that is critical to detect deliveries, are not captured by the baseline as frames are processed individually. For instance, a person getting close to a residence, bending towards the ground, and moving away is a strong indication for a delivery event. Therefore, having a model that accounts for temporal interactions across neighboring frames, creates the opportunity to learn enhanced representations. To this end, we propose to use a lightweight 3DCNN model that can process multiple frames at a time and extract rich temporal semantics. Fig. \ref{fig:proposed_pipeline} shows the overview of the proposed pipeline. 
For smart doorbell cameras, we need to have a mobile-friendly design that can be accommodated by the limited computational budget. To achieve this, we propose to use motion detection and tracking algorithm to reduce the spatial extent of frames to where motion occurs followed by a classifier to differentiate delivery from non-delivery events. Below, we discuss motion detection and tracking algorithm. Next, due to the computation-intensive nature of 3D convolutional and transformer based networks, we first discuss the lightweight networks that are mobile friendly. Based on this we set a 3D baseline and then design a novel semi-supervised attention module and adopt an evidential optimization objective to improve performance while preserving the computational complexity. 

\subsubsection{Motion Detection \& Tracking}
\label{subsubsec:motion}
\begin{figure}[t!]
    \centering
    \subfloat[][Full frame]{\includegraphics[width=.11\textwidth, height=.11\textwidth]{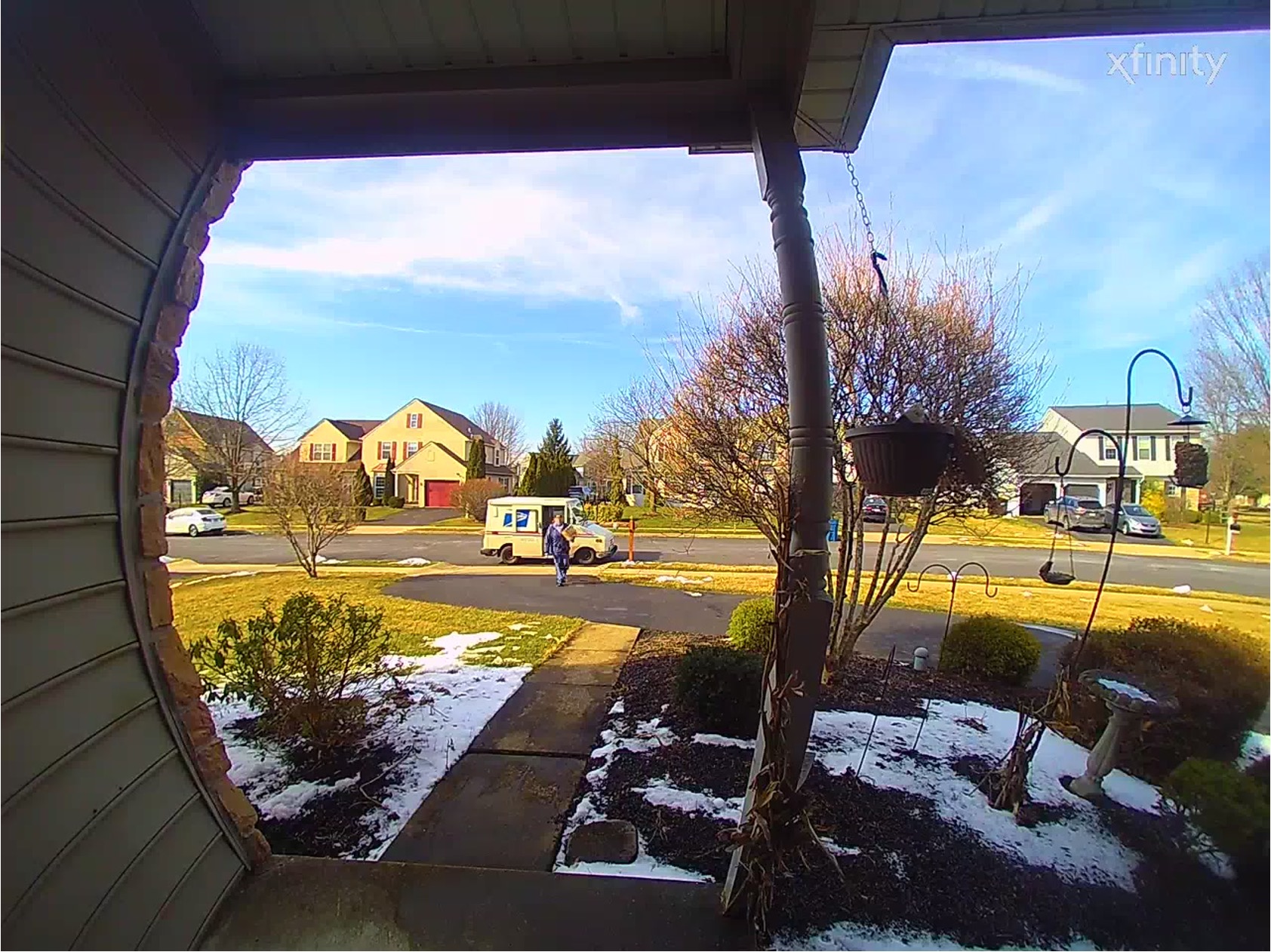}}~
    \subfloat[][Cropped]{\includegraphics[width=0.11\textwidth, height=.11\textwidth]{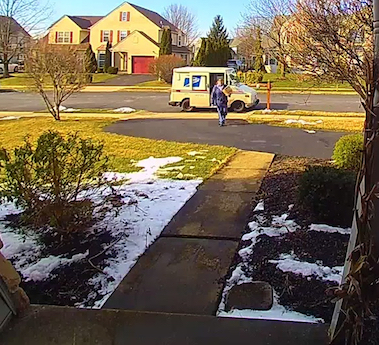}}~
    \subfloat[][Full frame]{\includegraphics[width=.11\textwidth, height=.11\textwidth]{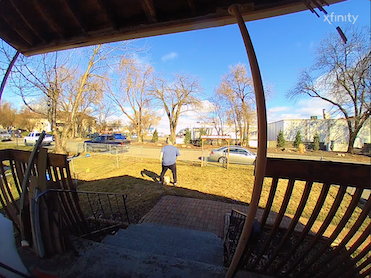}}~
    \subfloat[][Cropped]{\includegraphics[width=0.11\textwidth, height=.11\textwidth]{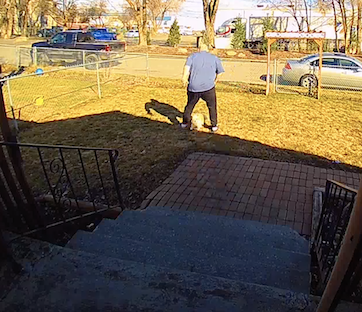}}\\
    \caption{Using motion to generate tighter activity proposals. (a) and (b) are full and cropped frames of a delivery event. (c) and (d) are full and cropped frames of a non-delivery event.}
    \label{fig:motion_cropped}
\end{figure}

To generate spatially-tighter proposals for delivery events compared to using full frames, we propose to use foreground motion to focus on regions that activities happen. This enables us to preserve a better pixel resolution as lightweight networks more often than not require small spatial size, \emph{e.g.} $112$x$112$. In case of using frames in their entirety, activity regions can only occupy a few number of pixels. Fig. \ref{fig:motion_cropped} shows the impact of using motion to generate tighter action proposals. 
Therefore, motion detection is an important pre-processing step to generate activity proposals. The algorithm for motion detection is based on Mixture of Gaussians (MOG) for background/foreground segmentation which adaptively models each pixel by a mixture of Gaussian distributions. This generates foreground motion mask containing motion blobs that are refined via adopting connected components. When a motion blob passes two thresholds, namely active time and variance, a motion event is triggered to signal the camera to query the scene. Active time indicates period of time in which a motion blob is continuously detected and tracked based on centroid distance measure. 
Variance criteria shows how much a blob has moved in the camera's field of view. This helps removing waving flags, leaves, and swaying trees which generate trivial motion events. Once a motion event is triggered, a thumbnail of fixed size is placed on the region where motion blobs reside. 

\subsubsection{Lightweight Backbone Architectures}
\begin{figure}[]
    \centering
    \includegraphics[width=0.3\textwidth]{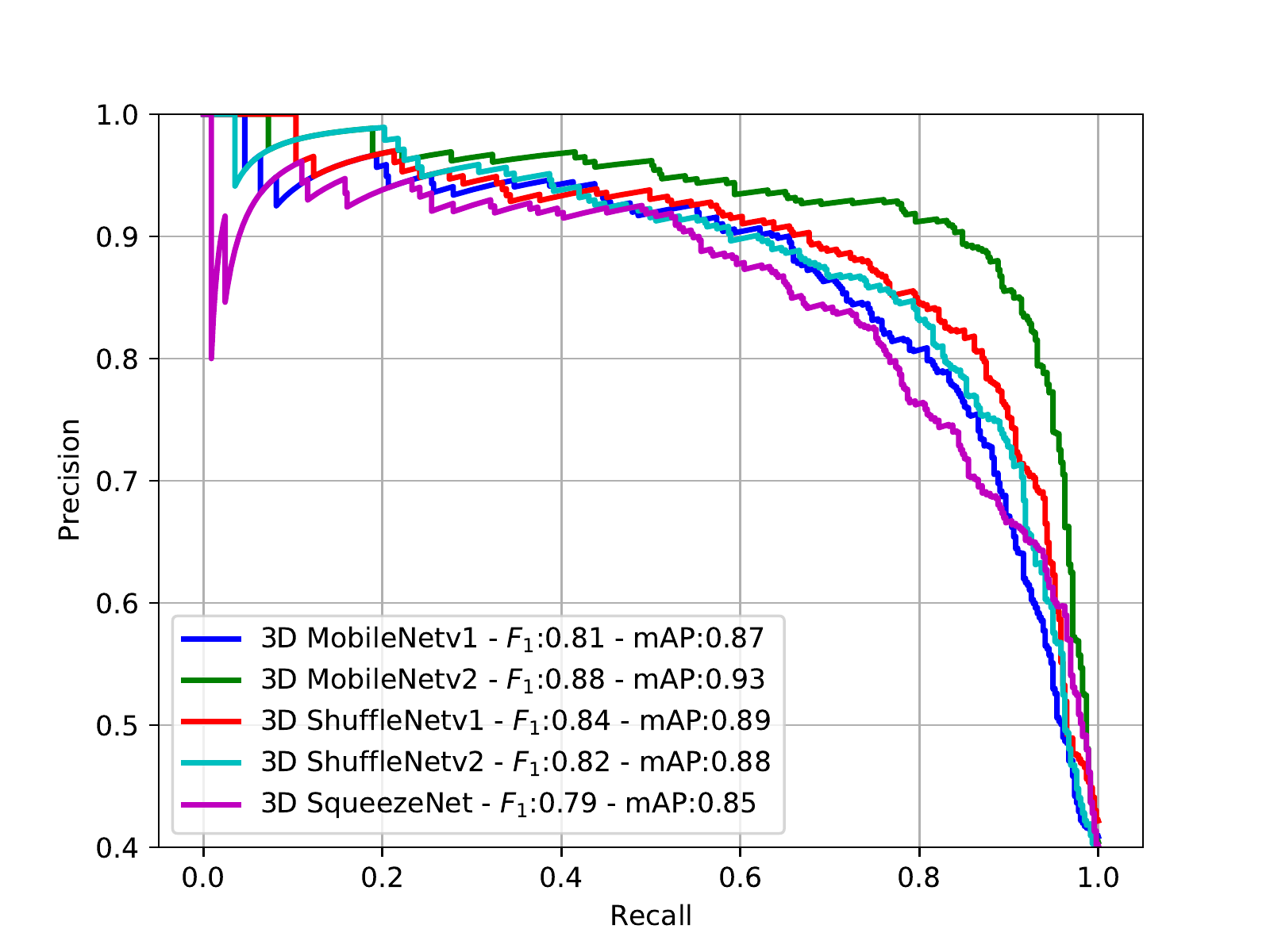}
    \caption{Precision-Recall comparison of lightweight 3DCNN architectures on the test set of our Doorbell delivery detection dataset discussed in section \ref{subsec:dataset}.}
    \label{fig:3dcnn_comp}
\end{figure}

As discussed in section \ref{sec:related_work}, 3D transformers are not particularly suited for mobile platforms to process visual data, hence we only focus on CNN based networks. Compared to their 2D counterparts, 3DCNNs have the ability to learn temporal interactions. This is achieved via additional parameters and higher computational complexity. Therefore, their adoption in resource-limited applications is constrained. To address this, \cite{kopuklu2019resource} introduced 3D versions of MobileNetv1 \cite{howard2017mobilenets}, MobileNetv2 \cite{sandler2018mobilenetv2}, ShuffleNetv1\cite{zhang2018shufflenet}, ShuffleNetv2\cite{ma2018shufflenet}, and SqueezeNet\cite{iandola2016squeezenet} which were developed for 2D mobile applications. Moreover, these networks are pre-trained on the Kinetics \cite{kay2017kinetics}, a large-scale human action recognition dataset, to provide robust weight initialization in the context of transfer learning when used in down-stream tasks. We use these networks as our candidate backbone architectures. We performed initial experiments on a subset of our curated dataset discussed in section \ref{sec:experiments}. Fig. \ref{fig:3dcnn_comp} compares the performance of these networks to distinguish delivery from non-delivery events in terms of Precision-Recall (PR) curve and $F_1$ score. Since MobileNetv2 obtains the highest accuracy, we base all our subsequent analysis on this network.
\subsubsection{Semi-supervised Attention}
\label{subsubsec:semi-attention}
\begin{figure}[t]
    \centering
    \includegraphics[width=0.45\textwidth]{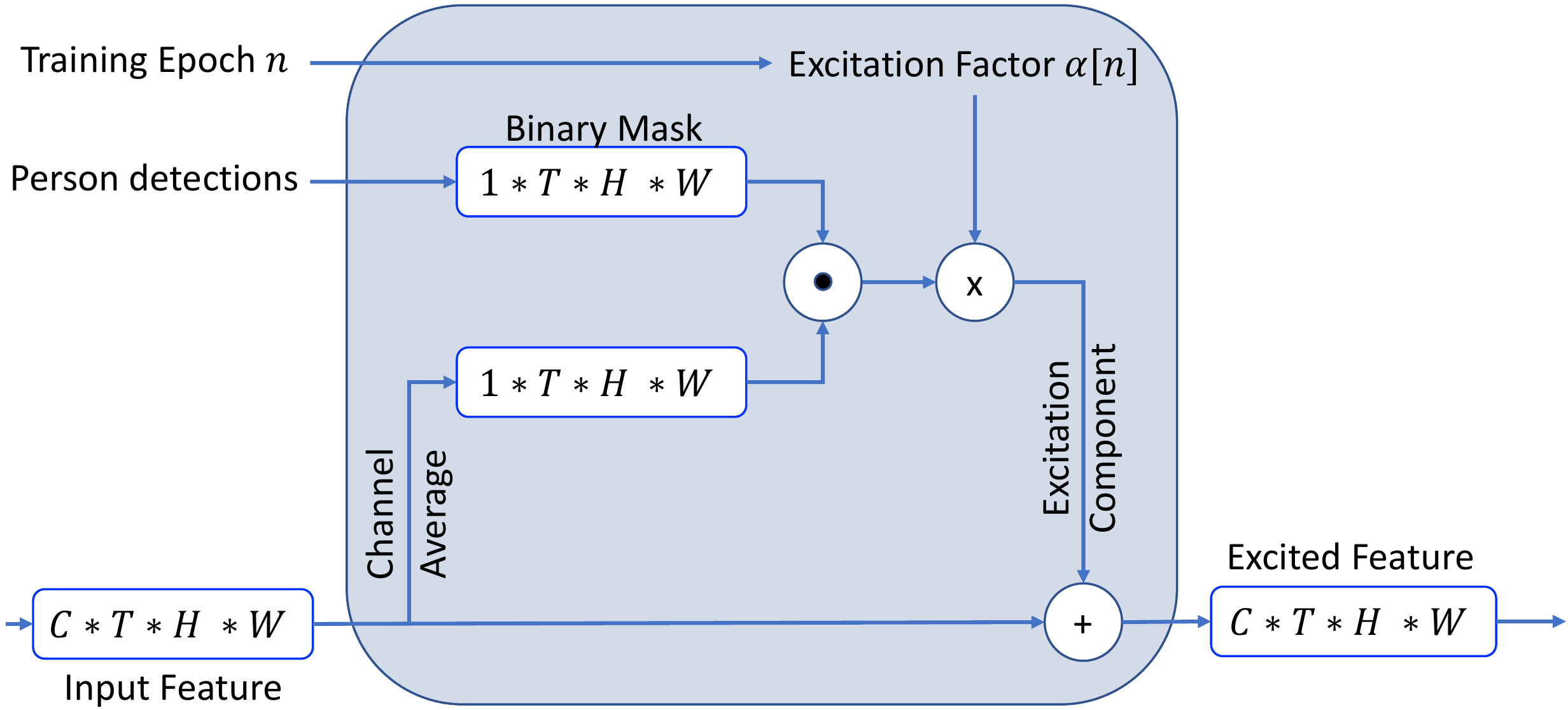}
    \caption{Excitation layer}
    \label{fig:how_to_excite}
\end{figure}
\begin{figure}
    \centering
    \includegraphics[width=0.48\textwidth]{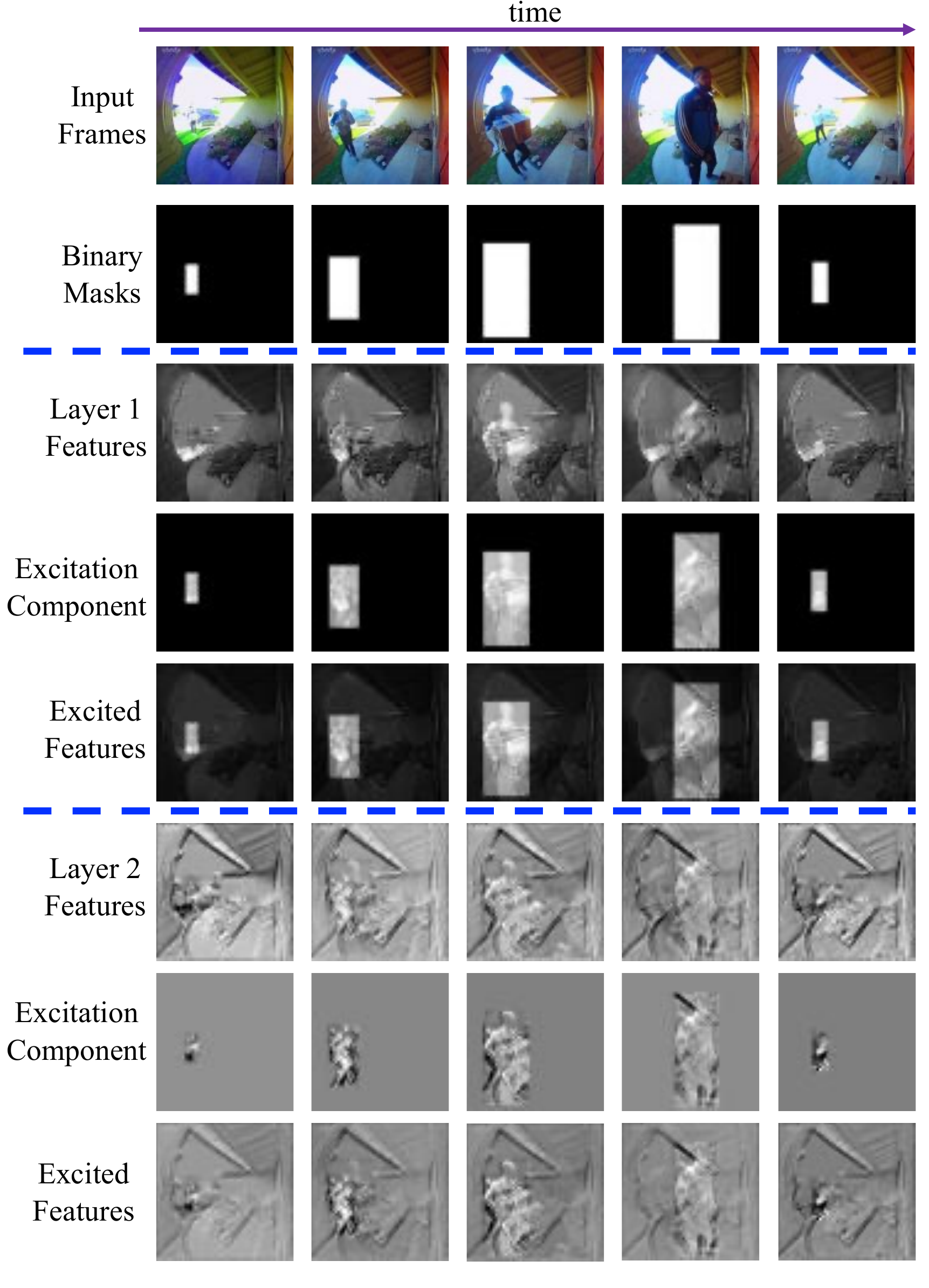}
    \caption{Exciting the first and second layer features of 3D MobileNetv2 for a sample video snippet in the first training epoch.}
    \label{fig:excitation_visualization}
\end{figure}

We are interested to investigate opportunities of improving accuracy without introducing additional compute and run-time complexity. Inspired by the works of \cite{derakhshani2019assisted,peri2020towards,Khorramshahi_2021_CVPR} which incorporate the paradigm of curriculum learning \cite{bengio2009curriculum} for object detection and vehicle analytics tasks, we devise a training mechanism to simplify the learning of delivery versus non-delivery events at early stages of optimization and gradually make the task more realistic as training progresses. The motivation for this simplification is that people can provide critical cues to distinguish deliveries. During training, we explicitly excite parts of the intermediate feature maps of the network that correspond to the location of people in the scene so that the network can better extract such signatures. However, as training progresses we gradually reduce our assistance, \emph{i.e.} the degree to which excitation happens, to let the network learn to performs enhanced feature extraction on its own. Consequently, this method of training can be viewed as a semi-supervised approach. Once training is concluded, excitation stops and the computational complexity will be the same as the original 3D MobileNetv2. To excite the output of the $l^{th}$ layer $f_l$ of shape $C * T * H * W$ where $C$, $T$, $H$, and $W$ represent number of channels, frames, height and width respectively, we generate $T$ binary single-channel masks of shape $H * W$ denoted by $m_l$ in which pixels corresponding to the bounding box location of people are set to one while the rest are set to zero. Afterwards, channel-averaged feature maps $\Tilde{f}_l = \sum_{c=1}^C f_l(c,.,.,.) / C$ are multiplied with $m_l$ in a point-wise manner. The resulting tensor is then multiplied by a scalar $\alpha[n]$ which is a function of training epoch $n$ as follows: $\alpha[n] = 0.5 * (1 + \cos (\pi n / N))$ where $N$ is the total number of epochs. The ensuing excitation component is finally added to the input feature maps. Fig. \ref{fig:how_to_excite} demonstrates the excitation operation and 
Equation \ref{eq:excite} expresses the mathematical relationship between the excited $f^{e}_l$ and original $f_l$ feature maps.
\begin{equation}
    \centering
    f^{e}_l = f_l + \alpha[n] * (\Tilde{f}_l . m_l)
    \label{eq:excite}
\end{equation}
For 3D MobileNetv2, we restricted the excitation to the outputs of first and second convolutional layers since they have the same temporal resolution as the input and the content of feature maps are not as abstract as deeper layers of the network. The impact of such excitation on the first and second layer features of the 3D MobileNetv2 is shown in Fig. \ref{fig:excitation_visualization}. Note that how regions of the feature map containing a person are highlighted with respect to the rest. This helps the network to focus more on the visual information of people and extract more robust representations in early stages of training.

\subsubsection{Evidence-based Delivery Detection}
\label{subsubsec:evidence}
To differentiate delivery from non-delivery events, a straightforward learning objective would be to obtain logits corresponding to each of the two classes followed by maximizing the likelihood of each input sample $p(y|x,\theta)$ where $x$, $y$, and $\theta$ are input sample, corresponding label, and model parameters. In practice this is achieved via Cross-Entropy loss which applies softmax function and minimizes the negative log-likelihood of the true class. While this approach is widely used, it that does not account for uncertainties when making predictions and only provides point estimates of class probabilities. In addition, the relative comparison of class probabilities cannot be used to quantify prediction uncertainty as softmax is known to inflate the probabilities \cite{guo2017calibration}. In contrast \cite{sensoy2018evidential} proposed a learning objective based on the theory of evidence \cite{dempster1968generalization} and subjective logic \cite{josang2016subjective} in which a predictor is tasked to gather evidence for any of the possible outcomes to formulate classification in conjunction with uncertainty modeling by considering a Dirichlet distribution as a prior on class probabilities. To realize this, an evidence function $g$ (which can be implemented as either ReLU, exponential, or soft-plus) is applied to the output of the network $h$ to ensure that outputs are always non-negative and are representative of the amount of evidence gathered by the network for each of the $K$ classes:
\begin{equation}
    \centering
    e_i = g(h_i(x;\theta)), \quad i = 1\dots K
\end{equation}
where $x$ and $K$ are the input video and the number of classes respectively. This is equivalent to gathering $K+1$ mass values $u$ and $b_i, i=1\dots K$ which are related through $u + \sum_{i=1}^{K}b_i = 1$. $u$ is the uncertainty of the prediction and $b_i$ is the belief mass corresponding to the $i^{th}$ class which is related to the evidence of $i^{th}$ class via $b_i = e_i/S$ where $S=\sum_{i=1}^{K}(\alpha_i)$ is referred to as the total strength of the Dirichlet distribution and $\alpha_i = e_i + 1, i=1\dots K$ are Dirichlet parameters. Based on this, uncertainty can be written as $u = K/S$ which is inversely proportional to the total strength $S$ or the total evidence gathered by the network $\sum_{i=1}^{K}e_i$. Therefore, gathering high evidence results in small uncertainty and vice-versa. Since class probabilities are assumed to follow Dirichlet distribution, \emph{i.e.} $\mathbf{p} \sim$ Dir($\mathbf{p}|\alpha)$ where $\mathbf{p} \in \mathbb{R}^{K}$, the average probability of the $i^{th}$ class can be computed as $\alpha_i/S$ \cite{sensoy2018evidential}. Therefore, the resulting loss function 
is computed via the following formulation:
\begin{equation}
\centering
\mathcal{L} = - \sum_{i=1}^{K}\mathbf{y}_i(\log(\alpha_i) - \log(S))    
\end{equation}
 $\mathbf{y}_i$ is the $i^{th}$ entry of the one-hot encoding label vector. Despite providing quantifying the uncertainty of predictions, this approach is deterministic and may suffer from the over-fitting caused by the training of neural networks. To alleviate this issue, a number of regularization terms are proposed. \cite{sensoy2018evidential} proposed a Kullback-Leibler (KL) term to encourage the network to generate zero evidence for a sample if it cannot be correctly classified. This is achieved by removing the generated evidence for the true class, \emph{i.e.} $\Tilde{\alpha}_i = \mathbf{y}_i + (1 - \mathbf{y}_i) . (e_i + 1)$, and minimizing the KL distance of the corresponding Dirichlet distribution Dir($\mathbf{p}|\Tilde{\alpha}_i$) from the one with zero total evidence, \emph{i.e.} $S = K$ and Dir($\mathbf{p}|\mathbf{1}$), which represents a uniform distribution. Note that $\mathbf{1}$ is the notation for all one vector. This KL term essentially discourages the network to over-fit and to generate evidence for samples about which it is uncertain:
\begin{equation}
    \centering
    \begin{split}
    \mathcal{L}_{KL} & = \log(\frac{\Gamma(\sum_{i=1}^{K}\Tilde{\alpha}_i)}{\Gamma(K)\prod_{i=1}^{K}\Gamma(\Tilde{\alpha}_i)}) \\
    & + \sum_{i=1}^{K}(\Tilde{\alpha_i}-1)\left(\psi(\Tilde{\alpha}_i)-\psi(\sum_{j=1}^{K}\Tilde{\alpha_j})\right)
    \end{split}
    \label{eq:KL_div}
\end{equation}
where $\Gamma(.)$ and $\psi(.)$ are gamma and logarithmic derivative of gamma function respectively. In addition to $\mathcal{L}_{KL}$ regularization, \cite{bao2021evidential} propose to calibrate the feature extraction network to be confident for its accurate predictions while being uncertain for it false predictions. To realize this goal, authors propose to maximize the Accuracy versus Uncertainty (AvU) utility function defined in \cite{krishnan2020improving}. AvU is formally defined as:
\begin{equation}
    \centering
    \text{AvU} = \frac{n_{AC} + n_{IU}}{n_{AC} + n_{AU} + n_{IC} + n_{IU}}
    \label{eq:avu_utility}
\end{equation}
In Eq. \ref{eq:avu_utility}, $n_{AC}$, $n_{AU}$, $n_{IC}$, and $n_{IU}$ are number of accurate and confident predictions, number of accurate and uncertain predictions, number of inaccurate and confident predictions, and number of inaccurate and uncertain predictions respectively. A well calibrated model should obtain high $n_{AC}$ and $n_{IU}$ and low $n_{IC}$ and $n_{AU}$. To regularize the learning of the model to achieve this objective, we draw inspiration from \cite{bao2021evidential} and add the following calibration objective to the overall loss function.  
\begin{equation}
    \begin{split}
        \mathcal{L}_{cal}  = & - \lambda_n \mathbbm{1}(\Tilde{y} = y ) p\log(1 - u) \\
        & - (1 - \lambda_n) \mathbbm{1}(\Tilde{y} \neq y)(1-p)\log(u)
    \end{split}
    \label{eq:calibration_loss}
\end{equation}
where $\hat{y} = \arg \max_{i} \{\alpha_i / S\}$, and $p = \max_{i}(\alpha_i / S)$ for $i\in\{1\dots K\}$ are the predicted class label and predicted probability for a given input sample. Note that $\mathbbm{1}(.)$ is the indicator function. Moreover, $\lambda_n$ is an epoch-dependent weight to adjust the contribution for each of the terms in the right hand side of Eq. \ref{eq:calibration_loss}. Specifically, $\lambda_n = \lambda_0e^{-\ln{(\lambda_0})n/N}$ is set to be exponentially-increasing ($\lambda_0 < 1$)  with respect to epoch index $n$. The intuition is that over the initial training epochs the model mainly makes inaccurate predictions and therefore it should be penalized to increase uncertainty for these predictions via $\mathbbm{1}(\Tilde{y} \neq y)(1-p)\log(u)$. However, as training progresses the model makes accurate predictions more often and therefore it should reduce the corresponding uncertainties which is enforced by $\mathbbm{1}(\Tilde{y} = y ) p\log(1 - u)$.
\section{Experiments} 
\label{sec:experiments}
Here we first describe the dataset we gathered and curated for our experiments. Afterwards the implementation details will be discussed followed by the presentation of the experimental results.

\subsection{Dataset}
\label{subsec:dataset}
\begin{figure}[]
    \centering
    \subfloat[][$0$ s]{\includegraphics[width=.09\textwidth]{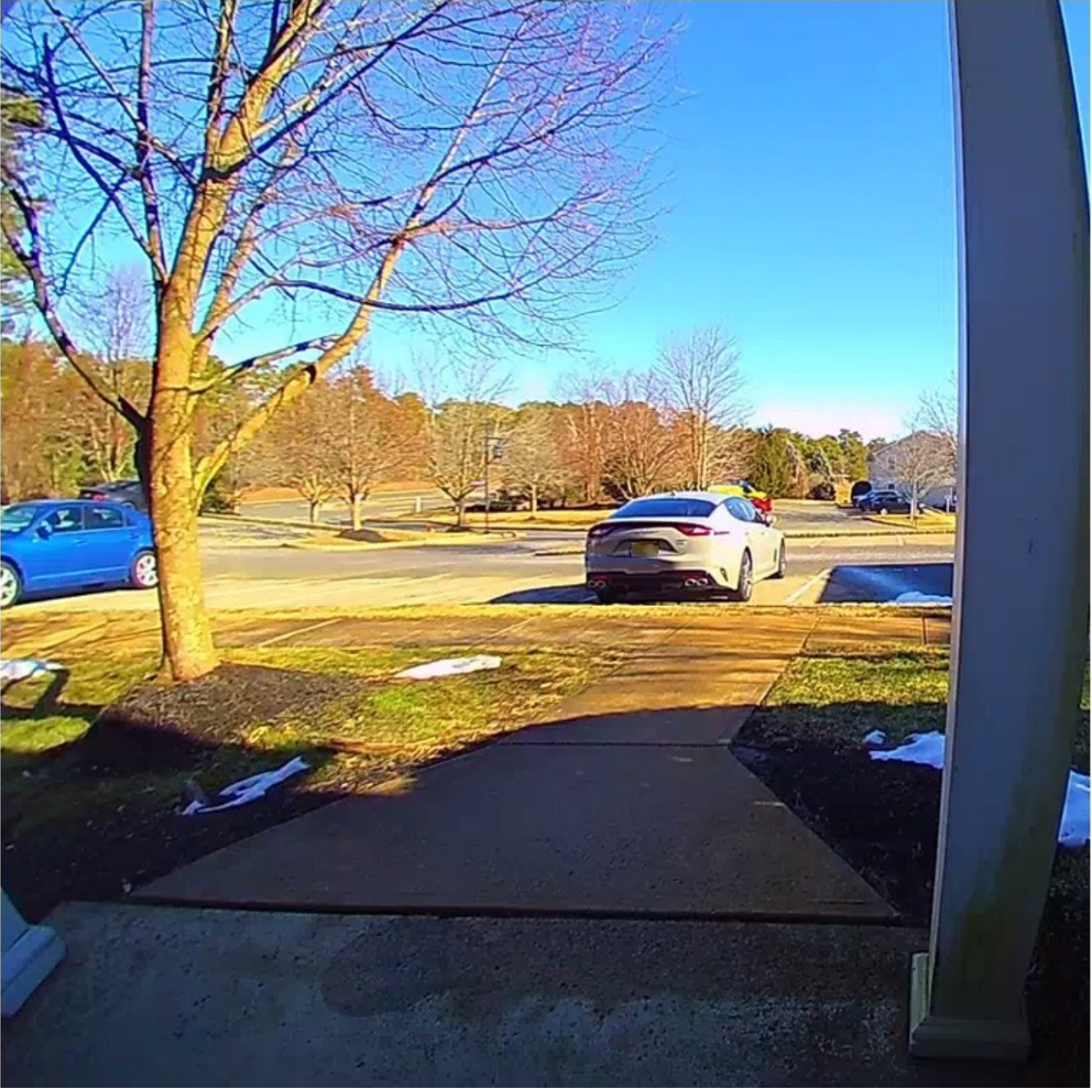}}~
    \subfloat[][$82$ s]{\includegraphics[width=.09\textwidth]{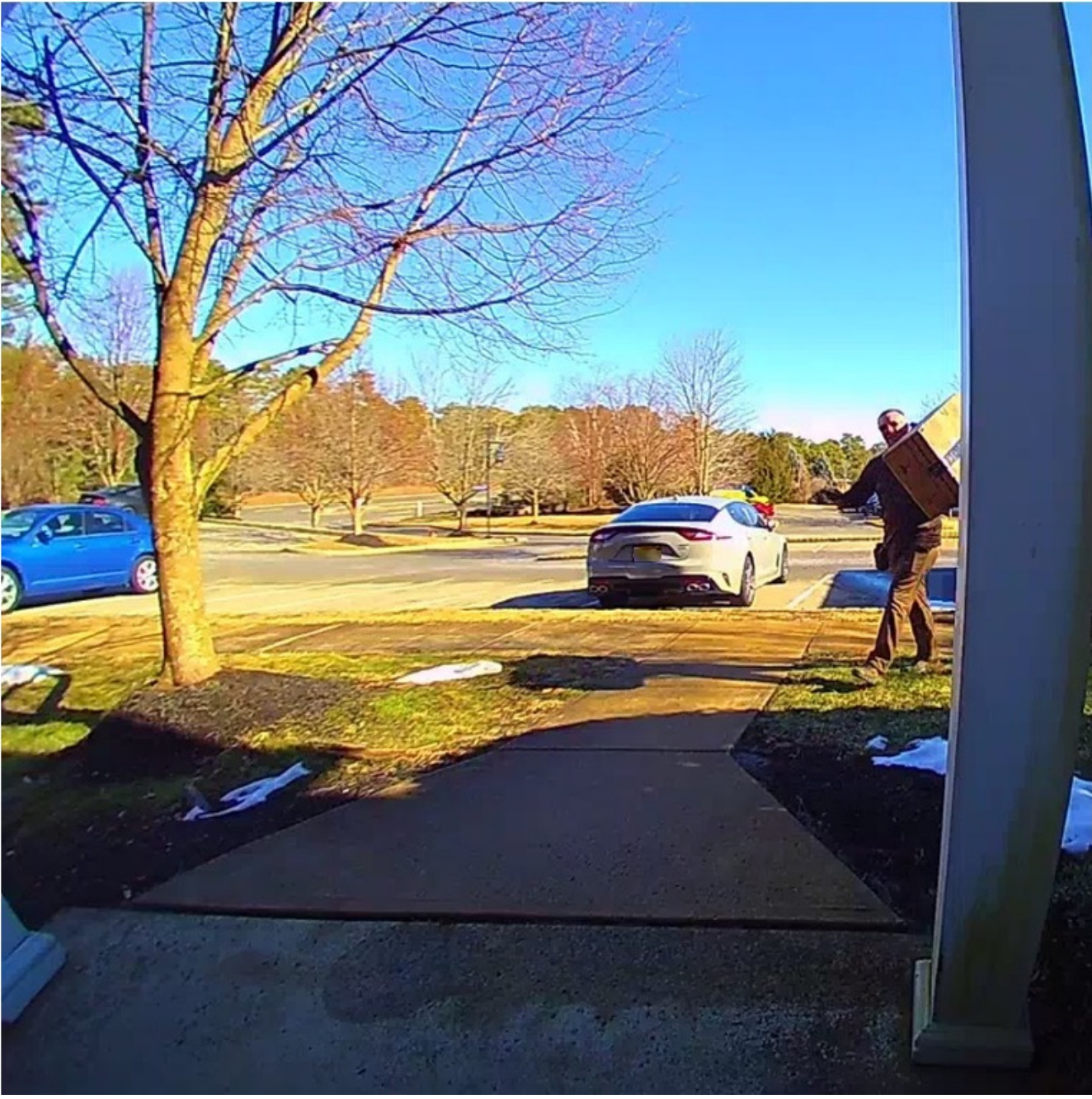}}~
    \subfloat[][$90$ s]{\includegraphics[width=.09\textwidth]{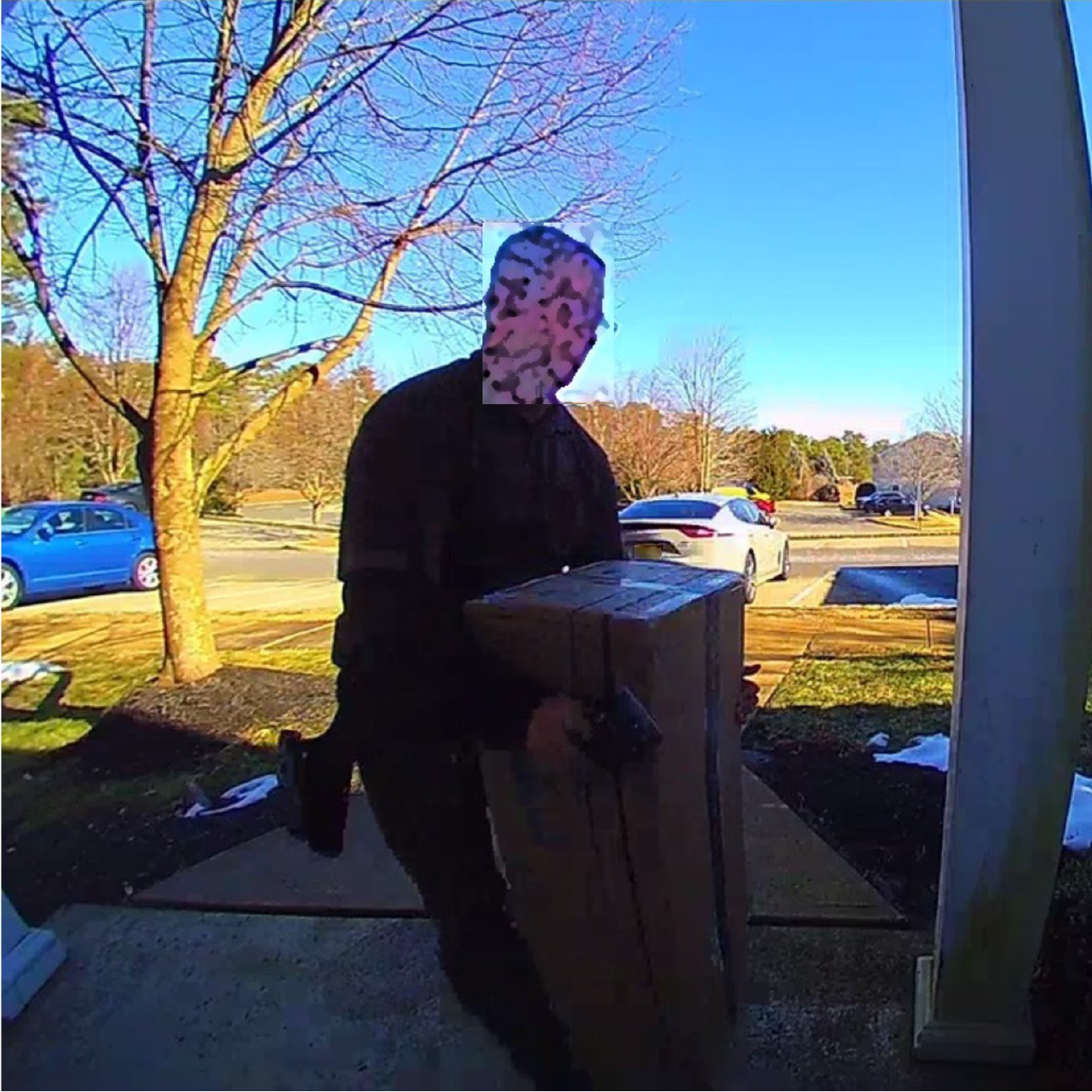}}~
    \subfloat[][$94$ s]{\includegraphics[width=.09\textwidth]{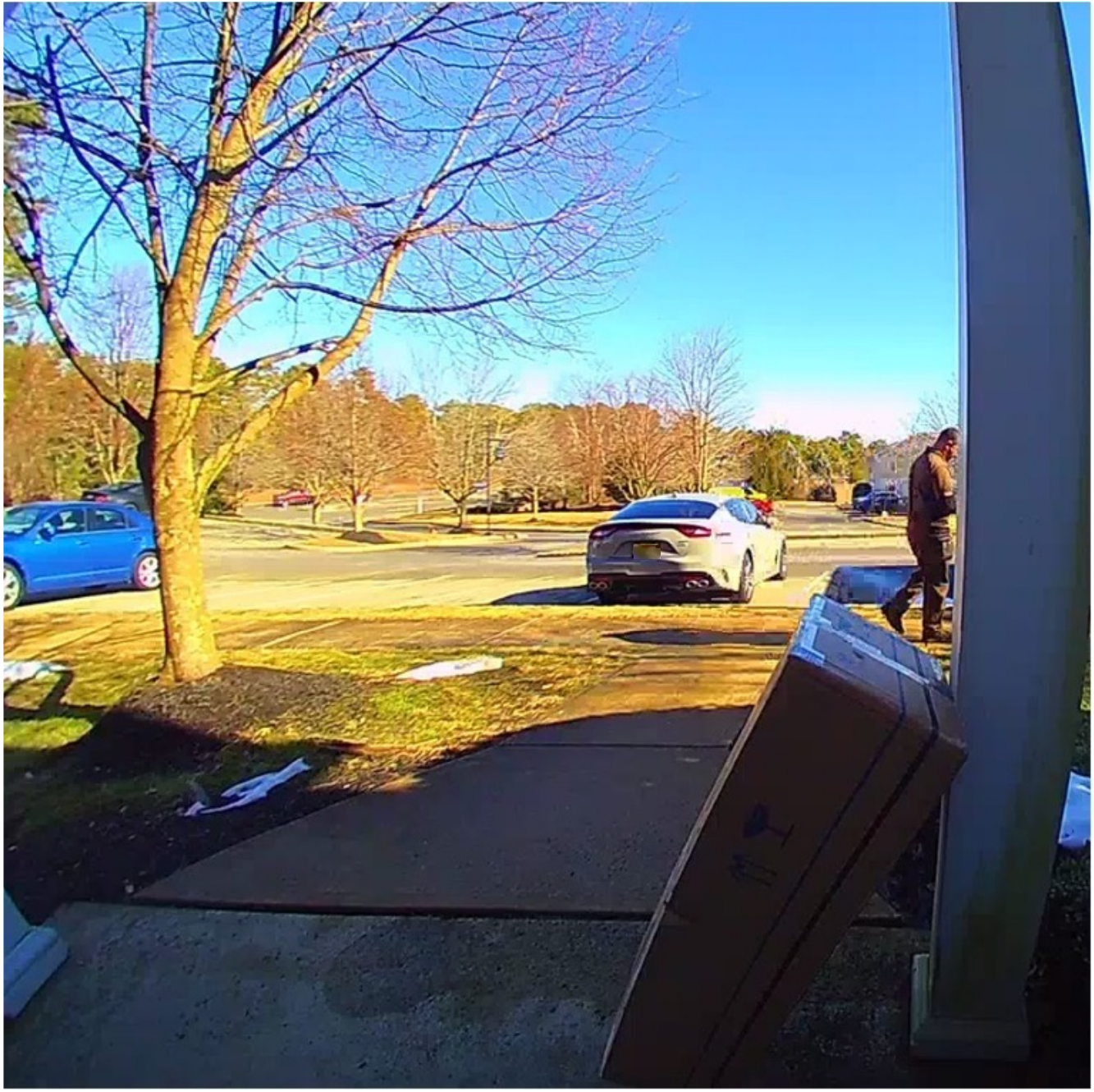}}~
    \subfloat[][$144$ s]{\includegraphics[width=.09\textwidth]{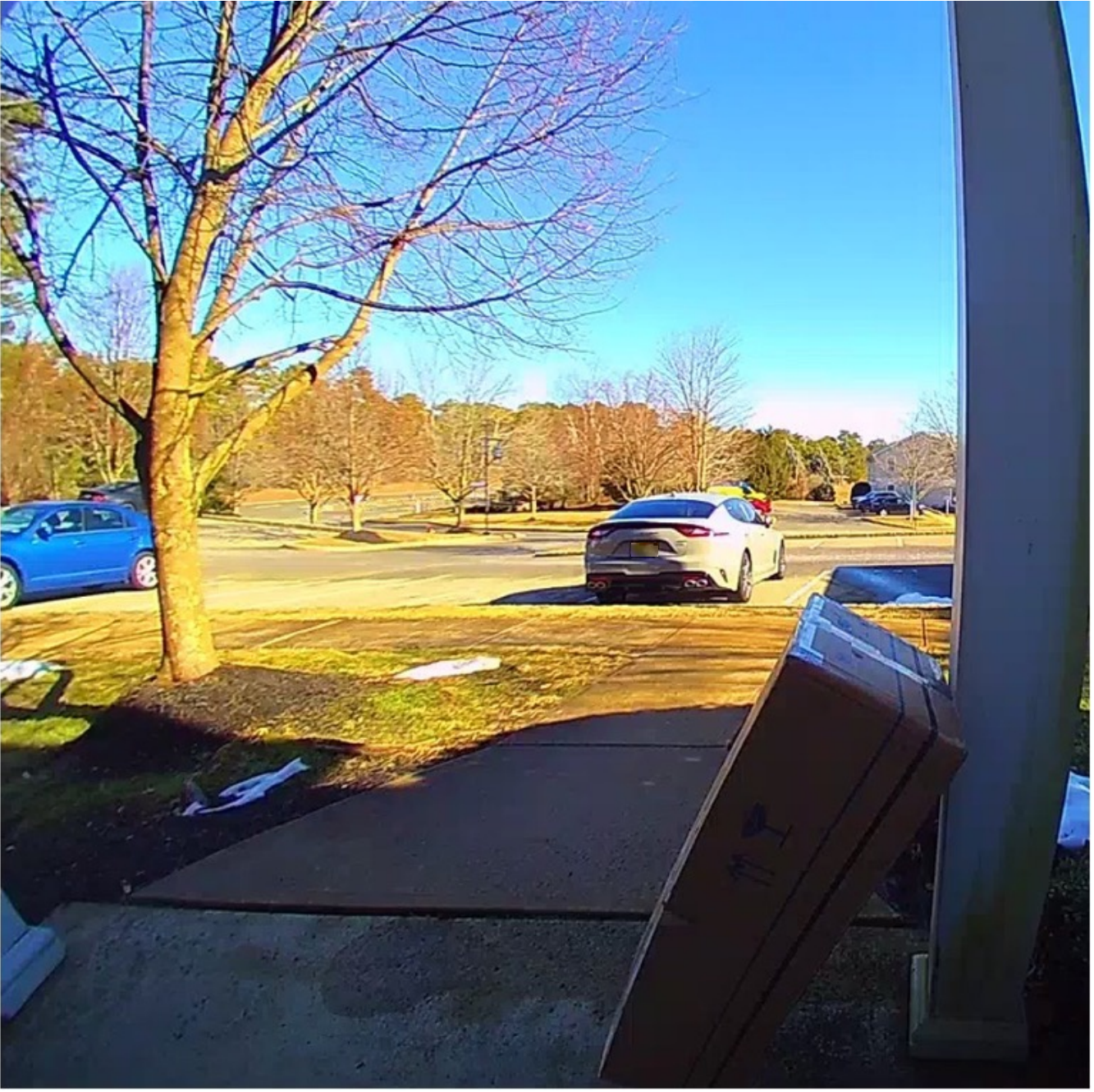}}
    \caption{Extent of deliveries are shorter than the length of videos. Here, in a $144$-seconds long video, delivery only lasts $12$ seconds.}
    \label{fig:delivery_extent}
\end{figure}

To the best of our knowledge, there are no publicly available dataset that is suited for our application. The closest dataset is UCF-Crime dataset \cite{sultani2018real} which includes only a handful of videos from security cameras installed at residential areas which cover a wide range of anomaly events that are not of interest for our study. Therefore, we choose to collect a video dataset by recording video snippets from static doorbell cameras installed at the front door of $339$ residences whose residents approved and signed the designated data collection agreement for this purpose. To ensure all videos contain activities, we started to record only when there was a motion trigger and stopped recording around two minutes after the start. This resulted in $5477$ videos with the resolution of $1280$x$960$. Sample frames of this dataset are shown in Fig. \ref{fig:sample_deliveries}. In the initial round of annotation process, all the videos were assigned a video-level tag to denote whether at least a delivery event happens during the entire duration of the video, which resulted in $1898$ delivery and $3579$ non-delivery video samples. However, we note that delivery events only occupy a small portion of the video as shown in Fig. \ref{fig:delivery_extent}. Therefore, to train our 3DCNN we require finer annotation for the start and end time of an instance within the video. Given a delivery event involves at least one person, to collect finer information, we run person detection and tracking on all videos to obtain person tracks. Afterwards we annotated person tracks with delivery/non-delivery tags. 
We used EfficientDet-D4 \cite{tan2020efficientdet} for detection and DeepSort \cite{Wojke2017simple,Bewley2016_sort} multi-object tracker with embeddings extracted by the bag of tricks for person re-identification \cite{luo2019bag} model with ResNet50\_IBN \cite{pan2018IBN-Net} backbone implemented in FastReID \cite{he2020fastreid} to compute tracks. This led to the collection of $2057$ person tracks delivering items and $2930$ person tracks that do not correspond to any delivery events, \emph{e.g.}, entering or exiting the residence, and playing in front lawn. Despite gathering fine annotations with tight spatial and temporal bounds, our delivery detection system is intended to be deployed on proposals generated based on motion events that are not tight around people in time and space and do not necessarily involve people, \emph{e.g.}, passing vehicles, presence of pets or wildlife. 
Therefore, we need to generate and label motion events to prepare our training and testing data. To generate these events we use the algorithm outlined in section \ref{subsubsec:motion}. To assign labels, labeled person tracks of delivery events are used to accelerate the process. To this end, we measured the overlap between a computed motion event and all person tracks within a video in terms of temporal and spatial Intersection over Union (IoU):
\begin{equation}
    \centering
    IoU_{t} = \frac{T(m_i \cap t_j)}{T(m_i \cup t_j)}, \qquad IoU_{s} = \frac{A(m_i \cap t_j)}{A(m_i \cup t_j)}
    \label{eq:iou}
\end{equation}
where $m_i$, $t_j$ are $i^{th}$ motion event and $j^{th}$ person track. In addition, $T(.)$, $A(.)$ are temporal length and the spatial area functions respectively. 
Note that initially we tried to use spatio-temporal (3D) IoU for measuring the overlap for a motion-track pair; however, we noticed that the overlap values become quite small even for closely matched pairs due to measuring the volume. Using two different thresholds, namely temporal $t_{min}$ and spatial $s_{min}$, provides more flexibility. 
Accordingly, $3625$ positive and $7248$ negative motion events were gathered and split into train, validation and test sets as reported by Table \ref{tab:dataset}.  
\begin{table}[]
    \centering
    \caption{Doorbell delivery detection dataset statistics. Note that cameras across splits are disjoint.}
    \resizebox{0.9\columnwidth}{!}{
    \begin{tabular}{c||c||c||c}
         \multicolumn{4}{c}{}\\
         \hline
         \multirow{2}{*}{Split} & \multirow{2}{*}{\# Cameras} & \multicolumn{1}{|c||}{\# Delivery} & \multicolumn{1}{|c}{\# non-Delivery} \\
         & & \multicolumn{1}{|c||}{Videos / Events} & \multicolumn{1}{|c}{Videos / Events} \\
         \hline
         \hline
         Train & $182$ & $1016$ / $2324$ & $1902$ / $3817$ \\
         Validation & $59$ & $416$ / $595$ & $769$ / $1680$ \\
         Test & $98$ & $466$ / $706$ & $900$ / $1751$ \\
         \hline
         Total & $339$ & $1898$ / $3625$ & $3579$ / $7248$ \\
         \hline
         \hline
    \end{tabular}}
    \label{tab:dataset}
\end{table}
\subsection{Implementation Details}
Each motion proposal is uniformly sampled by $16$ frames which are spatially resized to $112$x$112$ by preserving the original aspect ratio. During training, temporal jittering of $\pm10$ frames is applied to the start and end bounds of proposals as a form of data augmentation. Color jittering is also employed with the probability of $0.2$; brightness, contrast and saturation factors are uniformly chosen from $[0.9,1.1]$ while hue factor is uniformly selected from $[-0.1,0.1]$. Additionally, generated cuboids are horizontally flipped with the probability of $0.5$. To optimize Adam \cite{kingma2014adam} with decoupled weight decay AdamW \cite{loshchilov2017decoupled} is employed with gradient $L_2$ norm clipping at $0.25$ and the learning rate that is linearly warmed up to $5e-4$ in the first $5$ epochs and decayed at ${20}^{th}$ and ${40}^{th}$ epochs with gamma factor of $0.1$. Weight decay is set to $5e-4$ and the network is trained for the total of $50$ epochs. To alleviate the impact of easy samples, focal loss \cite{lin2017focal} with focusing parameter of $1.0$ is adopted.

\subsection{Evaluation Metric}
As our target task is to classify an input video as delivery or non-delivery, we use $F_1$ score and mean Average Precision (mAP) that presents the area under the precision-recall curve (AUC). Note that for a given video, multiple motion events may occur. In case the tag of the video is delivery, the classifier must classify at least one of those events as delivery and if the tag is non-delivery, the classifier must classify all the events as non-delivery. As we noticed high number of false positives with 2D baseline system \ref{sec:current_del_system}, we report the FPR as well.

\subsection{Experimental Results}
\begin{figure}[]
    \centering
    \includegraphics[width=0.3\textwidth]{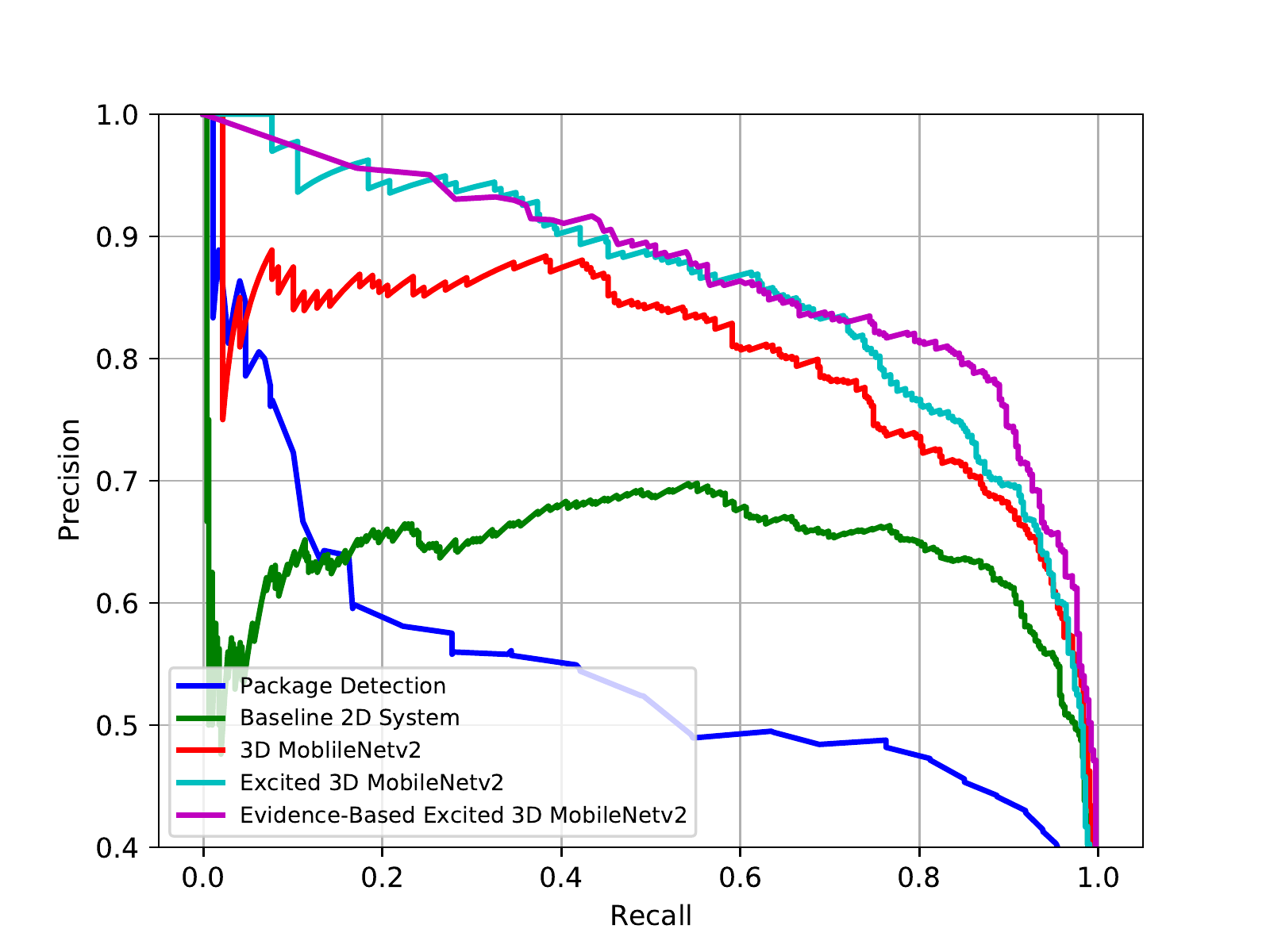}
    \caption{Precision-Recall comparison of package detection, baseline 2D pipeline and variants of 3D MobileNetv2 on the test set.}
   \label{fig:pr_comp}
\end{figure}
\begin{table}[]
    \centering
    \caption{Performance comparison for detecting delivery events on the test of Doorbell Delivery Detection dataset.}
    \resizebox{\columnwidth}{!}{
    \begin{tabular}{c||c|c|c|c}
         \cline{2-5}
         \multicolumn{1}{c||}{} & \multicolumn{4}{c}{Evaluation Metrics} \\
         \cline{1-5}
         \multirow{2}{*}{Model} & \multirow{2}{*}{$F_1$($\uparrow$)} & \multirow{2}{*}{mAP($\uparrow$)} & \multirow{2}{*}{FPR($\downarrow$)} &  Classification \\ 
         & & & & Accuracy (\%)($\uparrow$) \\
         \hline
         \hline
         Package Detection & $0.59$ & $0.54$ & $0.42$ & $63.74$ \\
         \hline
         2D Baseline & $0.73$ & $0.64$ & $0.24$ & $79.75$ \\
         \hline
         3D MobileNetv2 & $0.77$ & $0.80$ & $0.19$ & $83.02$ \\
         \hline
         Excited & \multirow{2}{*}{$0.79$} & \multirow{2}{*}{$0.85$} & \multirow{2}{*}{$0.15$} & \multirow{2}{*}{$85.05$} \\
         3D MobileNetv2 & & & & \\
         \hline
         Evidence-based  & \multirow{3}{*}{$\textbf{0.81}$} & \multirow{3}{*}{$\textbf{0.86}$} & \multirow{3}{*}{$\textbf{0.13}$} & \multirow{3}{*}{$\textbf{86.11}$} \\
         Excited 3D & & & & \\
         MobileNetv2 & & & & \\
         \hline
         \hline
    \end{tabular}
    }
    \label{tab:models_comp}
\end{table}
This section compares the 2D baseline system described in section \ref{sec:current_del_system}, the 3D MobileNetv2, the 3D MobileNetv2 with semi-supervised attention module outlined in section \ref{subsubsec:semi-attention}, namely excited MobileNetv2, and the excited 3D MobileNetv2 optimized with the evidence-based objective of section \ref{subsubsec:evidence}. Additionally, as package detection is offered in many smart home solutions, we choose to evaluate its applicability for delivery detection task. 
To this end, we train a COCO \cite{lin2014microsoft} pre-trained MobileNet-SSD v2 object detector on $4405$ manually labeled images with package annotations. Fig. \ref{fig:pr_comp} plots the precision-recall curves and Table \ref{tab:models_comp} reports evaluation metrics for each of these models when evaluated on the test set. Unsurprisingly, package detection has a significantly inferior performance as detecting small and occluded packages is challenging. Also drastic variation in size, shape and appearance of delivered items further exacerbates this performance gap. Therefore, solutions based on package detection are not suited for detecting delivery instances. From Table \ref{tab:models_comp} it is also seen that a 2D model compared to 3D alternatives performs at a lower level due to inability of modeling temporal interactions. 
\begin{table}
    \caption{Efficiency comparison between baseline 2D system and the proposed system based on 3D MobileNetv2 architecture.}
    \centering
    \label{tab:efficiency}
    \resizebox{\columnwidth}{!}{
    \begin{tabular}{c||c|c|c|c}
         \hline
         \multirow{2}{*}{System} & \multirow{2}{*}{Input} & \multicolumn{1}{|c}{Inference time} & 
         \multicolumn{1}{|c}{Binary Size} & \multicolumn{1}{|c}{FLOPS} \\ 
         & & \multicolumn{1}{|c}{(ms/input)} & \multicolumn{1}{|c}{(MB)} & \multicolumn{1}{|c}{(G)}\\
         \hline \hline
         \multirow{1}{*}{2D Baseline} &\multicolumn{1}{c|}{$1$x$1280$x$960$} & \multirow{1}{*}{$44.5$} & \multirow{1}{*}{$7.0$} & 
         \multirow{1}{*}{$1.41$} \\
         \hline
         \multirow{1}{*}{3D MobileNetv2} & \multicolumn{1}{c|}{$16$x$112$x$112$} & \multirow{1}{*}{$69.94$} & \multirow{1}{*}{$2.9$} &
         \multirow{1}{*}{$0.55$} \\ 
         \hline
         \hline
    \end{tabular}
    }
\end{table}
We observe the increase of $4\%$ and $16\%$ in $F_1$ and mAP scores when we incorporate the temporal information via 3D MobileNetv2 which also results in reducing the FPR by $5$ points generating much fewer false delivery notifications. Additionally, training the 3D MobileNetv2 with our novel semi-supervised attention module and the incorporation of evidence-based optimization objectives increases $F_1$ and mAP scores by $5.1\%$ and $7.5\%$ and reduced the FPR by $31\%$. These novelties further enhances the performance of 3D MobileNet in a meaningful manner without introducing any additional overhead during test time. This is of great importance for a resource limited design to maintain a fixed computational budget during inference. 

\begin{figure}[]
    \centering
    \subfloat[][]{\includegraphics[width=.09\textwidth]{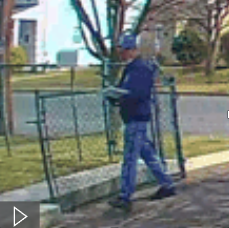}}~
    \subfloat[][]{\includegraphics[width=.09\textwidth]{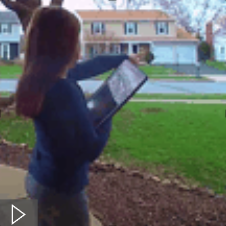}}~
    \subfloat[][]{\includegraphics[width=.09\textwidth]{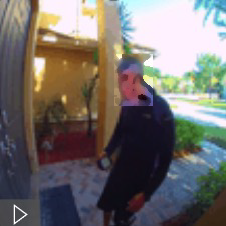}}~
    \subfloat[][]{\includegraphics[width=.09\textwidth]{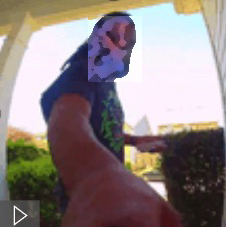}}~
    \subfloat[][]{\includegraphics[width=.09\textwidth]{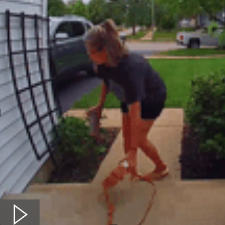}}~
    \caption{Uncertain predictions by our network.}
    \label{fig:uncertain_samples}
\end{figure}

It is also important to compare the 2D baseline model with the 3D MobileNetv2 in terms of run-time speed, binary size of quantized models and the number of floating point operations (FLOPS) which are critical when deployed on an edge device. Note that Excited 3D MobileNetv2 and Evidence-based Excited 3D MobileNetv2 share the same run-time, binary size and FLOPS as 3D MobileNetv2 during inference. Table \ref{tab:efficiency} provides this comparison. While the inference time of 2D baseline system is $36\%$ lower, we have to note that the system continuously queries the scene at a fixed rate and performs $2.5$ times more operations ($1.41$ GFLOPS) per forward pass compared to 3D MobileNetv2 ($0.55$ GFLOPS) which performs inference once a motion event is concluded. Also the required memory to store 3D MobileNetv2 is much smaller compared to the 2D baseline which provides opportunities for increasing the complexity and potentially the accuracy of a prospective model. Finally we would like to highlight an additional benefit of using an evidence-based objective compared to Cross-Entropy. We can compute the average uncertainty score for the samples within the validation set on which the model made mistakes. We further use this value as a threshold to remove the predictions whose uncertainty values are higher when processing the test set as presented in Table \ref{tab:uncertainty_removal}.
\begin{table}[]
    \centering
    \caption{Removing samples with uncertainty score above $0.16$ computed from the validation set. Total samples removed from test set: $89$ videos.}
    \resizebox{1\columnwidth}{!}{
    \begin{tabular}{c||c|c|c|c}
         \cline{2-5}
         \multicolumn{1}{c||}{} & \multicolumn{4}{c}{Evaluation Metrics} \\
         \cline{1-5}
         \multirow{2}{*}{Model} & \multirow{2}{*}{$F_1$($\uparrow$)} & \multirow{2}{*}{mAP($\uparrow$)} & \multirow{2}{*}{FPR($\downarrow$)} &  \multicolumn{1}{c}{Classification} \\ 
         & & & & \multicolumn{1}{c}{Accuracy (\%)($\uparrow$)} \\
         \hline
         \hline
         Evidence-based & \multirow{2}{*}{$0.81$} & \multirow{2}{*}{$0.86$} & \multirow{2}{*}{$0.13$} & \multirow{2}{*}{$86.11$} \\
         Excited 3D MobileNetv2 & & & & \\
         \hline
         Evidence-based  & \multirow{3}{*}{$\textbf{0.83}$} & \multirow{3}{*}{$\textbf{0.87}$} & \multirow{3}{*}{$\textbf{0.12}$} & \multirow{3}{*}{$\textbf{87.92}$} \\
         Excited 3D MobileNetv2 & & & & \\
         + Uncertainty Removal & & & & \\
         \hline
         \hline
    \end{tabular}
    }
    \label{tab:uncertainty_removal}
\end{table}
By applying this threshold, we remove $89$ videos from our test set which not only increases $F_1$ and mAP scores, but also reduces the FPR. Here we visualize randomly selected samples about which the model was uncertain in Fig. \ref{fig:uncertain_samples}. 
It is seen that, these sample have flavors of delivery events. For instance, in (a) a mailman is going towards a neighboring house, in (b) a family member is holding a tablet which is what most delivery personnel do after delivering an item to submit proof of delivery, in (c) a mail man is picking an item from front door for either shipping or returning, in (d) a food delivery person with no uniform is seen, and in (e) the resident is putting down her belongings at the front door. Therefore, uncertainty score can be used effectively to reduce the number of false predictions.

\section{Conclusion} 
\label{sec:conclusion}
In this work we presented a mobile-friendly pipeline to perform the task of delivery detection in contrast to package detection on resource-limited platforms such as doorbell cameras. The proposed system has the capacity of modeling temporal interactions in video streams to enhance its predictions over a 2D model. In addition, we have improved the accuracy of our designed system by a considerably through the novel incorporation of a semi-supervised attention module, namely excitation layer. We have also benefited from the advances in the theory of evidence and subjective logic to modify the optimization objective of the system. This not only boosts the system performance but also quantifies the uncertainty of predictions made by the network and provides the opportunity to enforce a minimum level of certainty to further advance predictions. We emphasize that all these improvements are achieved without adding any inference-time computation and memory overhead to the proposed design.

{\small
\bibliographystyle{ieee_fullname}
\bibliography{egbib}
}

\end{document}